# *Advanced Meta-Ensemble Machine Learning Models for Early and Accurate Sepsis Prediction to Improve Patient Outcomes*

MohammadAmin Ansari Khoushabar[1], Parviz Ghafariasl[2]

[1] Department of Industrial Engineering, Science and Research Branch

Islamic Azad University, Tehran, Iran

amin.ansari@srbiau.ac.ir

[2] Department of Industrial and Manufacturing Systems Engineering

Kansas State University, Manhattan, KS, USA

Parvizghafari@ksu.edu

**Abstract:**

Sepsis, a critical condition resulting from the body's response to infection, poses a major global health crisis affecting all age groups. Timely detection and intervention are crucial for reducing healthcare expenses and improving patient outcomes. This paper examines the limitations of traditional sepsis screening tools like Systemic Inflammatory Response Syndrome (SIRS), Modified Early Warning Score (MEWS), and Quick Sequential Organ Failure Assessment (qSOFA), highlighting the need for advanced approaches. We propose using machine learning techniques—Random Forest, XGBoost, and Decision Tree models—to predict sepsis onset. Our study evaluates these models individually and in a combined meta-ensemble approach using key metrics such as Accuracy, Precision, Recall, F1 score, and Area Under the Receiver Operating Characteristic Curve (AUC-ROC). Results show that the meta-ensemble model outperforms individual models, achieving an AUC-ROC score of 0.96, indicating superior predictive accuracy for early sepsis detection. The Random Forest model also performs well with an AUC-ROC score of 0.95, while XGBoost and Decision Tree models score 0.94 and 0.90, respectively. This research highlights the potential of machine learning, particularly the meta-ensemble approach, to enhance early sepsis prediction. By leveraging the strengths of multiple models, this approach provides a robust tool for improving patient outcomes, demonstrating the transformative potential of machine learning in healthcare.

## 1-Introduction

Sepsis, a fatal condition brought on by the body's reaction to infection, is a leading cause of death globally and has turned into an epidemiological emergency. All age groups can get sepsis, which increases mortality rates. For instance, sepsis affects about 1.7 million persons in the United States, and complications from sepsis result in nearly 270,000 fatalities each year [1]. Sepsis is also the most costly disease in all of the states of America, costing $24 billion in U.S. hospital expenses in 2013 [2]. Sepsis has complex pathophysiological pathways that result in various signs and symptoms that are not always obvious [3]. As some patients are already in a septic state during admission, recent studies show that delaying the necessary antibiotic treatment increases the death rate per hour. Early risk factor detection and appropriate monitoring before the onset of clinical symptoms significantly impact overall mortality rates and the cost burden of sepsis [4].

Systemic Inflammatory Response Syndrome (SIRS), Modified Early Warning Score (MEWS), rapid Sequential Organ Failure Assessment (qSOFA), and other sepsis screening techniques are currently ineffective at detecting sepsis [5]. Sepsis can develop into severe sepsis and septic shock without prompt and effective treatment, which increases the fatality rate [6].

Numerous studies show that early sepsis prediction permits early intervention and significantly enhances patient outcomes [7,8]. The early identification of sepsis is complicated because common indications and symptoms, such as fever, chills, fast breathing, and high heart rate, mirror symptoms of other illnesses. Additionally, it is practically impractical to forecast sepsis a few minutes before it manifests, even with excellent accuracy. Therefore, a successful prediction model should generate more powerful signals as the event draws near, providing timely alerts. An effective screening tool could aid medical practitioners in concentrating their efforts on patients who are at a high risk of going into septic shock, hence reducing this issue. No specific clinical tool for septic shock screening has been created, though. Rapid response teams frequently employ the Quick Sequential Organ Failure Assessment (qSOFA), initially developed for sepsis screening or the Modified Early Warning Score (MEWS), to diagnose and evaluate patient severity. It may also be helpful for septic shock screening [7]. However, these techniques do not consider potential interactions between the factors and only use a small number of variables, which is not ideal for septic shock screening.

Healthcare settings are investigating machine learning (ML) methods for various purposes [9,10, 30]. ML systems can precisely diagnose sepsis using accumulated data from intensive care units (ICUs) and emergency rooms [11, 12]. Many researchers have concentrated on ML techniques to outperform severity scoring systems and attain high accuracy. Machine learning tries to produce algorithms that learn and build models for data analysis and prediction, delivering quick results [13]. Based on PhysioNet data gathered from three hospitals, the current study aims to use machine learning techniques, such as XGBoost, Random Forest, and Decision Tree, to detect sepsis in patients throughout their ICU stay. But how can we make these machine learning models more accurate and reliable? We look into using a meta-ensemble approach to get around these problems and improve accuracy. An approach to machine learning called ensemble learning aims to improve predictive performance by mixing predictions from various models. Lowering the generalization error in predictions is the aim of employing ensemble models [29]. The ensemble approach reduces model prediction error when the basis models are diversified and independent. The approach predicts by adding each output individually. Even though the ensemble model comprises multiple base models, it behaves and operates like a single model [22, 31]. We expect our meta-ensemble approach to perform better than the individual base models, advancing predictive modeling for early sepsis predictions.

**Related Work**

In the realm of predicting sepsis, a variety of models and approaches have been explored in previous research.

Abromavičius et al. [18] addressed the problem of early sepsis detection using data from the PhysioNet/Computing in Cardiology Challenge 2019. They developed predictive models using machine learning algorithms like Decision Tree, SVM, Naive Bayes, and Ensemble methods. Twenty-four features were extracted from the ICU dataset's vital signs, lab values, and demographics. Features included means, standard deviations, and max-min differences of vital signs. The imbalanced dataset (only 2% sepsis cases) was handled by adjusting the classification cost function to weigh sepsis cases higher. Two separate ensemble models were proposed based on the length of ICU stay - one for <56 hrs and one for >56 hrs. The best model used AdaBoost with decision trees for <56 hrs of data and achieved a 0.242 Utility score on the test set. For >56 hours of data, an ensemble method called discriminant subspace classification was used, which earned a 0.245 Utility score—combining these two models in a staged approach based on ICU length of stay improved performance over individual models. The authors analyzed performance metrics like AUCROC, precision, etc., and found that the utility score was most appropriate for this imbalanced dataset. In summary, the key contributions were using ICU length of stay to develop separate models, adjusting classification cost for imbalance, and comparing various metrics to show that the utility score was most suitable. The staged modeling approach improved sepsis prediction.

Barton et al. [19] developed a gradient-boosted decision tree model to predict sepsis onset up to 48 hours before. The model only used six vital signs as input - heart rate, respiration rate, temperature, SpO2, systolic BP, and diastolic BP. Retrospective data from two hospitals, UCSF and Beth Israel Deaconess Medical Center, was utilized. The model achieved AUROC of 0.88 at sepsis onset, 0.84 at 24 hours prior, and 0.83 at 48 hours prior. This outperformed standard rules-based scores like SIRS, SOFA, MEWS, and qSOFA at the onset time. The model maintained performance when trained on UCSF data and tested on BIDMC data. Key predictors were age, temperature, heart rate, and blood pressure trends. The authors conclude the model can predict sepsis risk up to 48 hours in advance using just vital signs. Limitations include retrospective design and class imbalance in longer prediction windows. In summary, this study demonstrated the good performance of a machine learning model for early sepsis prediction using only commonly available vital signs, with the ability to predict up to 48 hours before onset. The model outperformed existing scoring systems.

Using ICU data, Kaji et al. [20] developed models to predict sepsis, myocardial infarction (MI), and the need for vancomycin. A large dataset containing 119 variables, including vital signs, lab values, medications, etc., from over 90,000 patients was used. Recurrent neural networks (RNN) with long short-term memory (LSTM) units and attention mechanisms were implemented. The attention mechanism allowed the model to learn the most relevant features at each timestep. Separate models were built for each prediction task (sepsis, MI, vancomycin) using Keras and TensorFlow. The sepsis model achieved an AUC of 0.876, the MI model's AUC was 0.823, and the vancomycin model got an AUC of 0.833 on the test set. Attention analysis revealed dynamic changes in feature importance over time towards the onset of the conditions. Interpretability was improved by using attention weights to identify influential features at each timestep.

In summary, this study utilized deep learning approaches to make clinical predictions from EHR data, reaching high accuracy. The attention mechanism provided some model interpretability. An extensive set of variables over long time periods was used. The fundamental limitations were model complexity and reliance on large datasets. Overall, the study demonstrated promising capabilities of deep learning for clinical event prediction in ICU settings.

Bloch et al. [21] developed a novel feature extraction approach, focusing on the hypothesis that unstable patients are more likely to develop sepsis during their ICU stay. The study used electronic medical records (EMRs) of patients admitted to the ICU at the Rabin Medical Center in Israel from 2007 to 2014. These EMRs contained real-time clinical and laboratory data, drug administration, and medical notes. The authors extracted features from four common vital signs: mean arterial blood pressure, heart rate, respiratory rate, and temperature. They analyzed the variability in these vital signs, especially focusing on the interval leading up to a prediction moment for sepsis onset. Five different machine learning algorithms were implemented using R software packages. These algorithms aimed to calculate a patient's probability of developing sepsis within the next 4 hours based on the vital sign data from the previous 8 hours. The algorithms included logistic regression, support vector machines (SVM) with different kernels, and artificial neural networks (ANNs). The performance of the models was evaluated using various metrics, including the area under the ROC curve (AUC), sensitivity, specificity, accuracy, negative predictive value, and positive predictive value. The SVM with a radial basis function (SVM-RBF) achieved the highest AUC of 88.38%, indicating its predictive solid performance. This model also had high positive predictive value and specificity. The authors found that predicting sepsis 4 hours in advance was a suitable balance between accuracy and practicality. The authors compared their model to previous approaches that used different sets of features and data collection periods, demonstrating the superiority of their model in predicting sepsis onset. The paper presents a promising approach to early sepsis detection using readily available vital sign data, which could be applied in ICUs to improve patient outcomes.

Zeyu Liu et al. [23] proposed a two-stage framework called MLePOMDP combining machine learning (ML) and a partially observable Markov decision process (POMDP). In the first stage, ML models like random forests and neural networks make real-time predictions from physiological data. In the second stage, these ML predictions are fed as "observations" to a POMDP model. The POMDP models the underlying sepsis progression and makes optimal decisions on when to predict sepsis. Three actions are defined: sepsis, non-sepsis, and wait-and-see. The wait-and-see action acts as a buffer to avoid false alarms. Analytically, the optimal policy is proven to have a threshold structure. This allows efficient solution of the POMDP. On an ICU dataset, MLePOMDP improves precision by up to 9% and reduces false alarms by 28% compared to ML benchmarks. Testing on an external eICU dataset shows the framework generalizes well. In summary, the key contributions are the proposed ML-POMDP framework, analytical results on structural properties, and improved precision and false alarm reduction for early sepsis detection.

Lauritsen et al. [24] developed a deep learning model for early detection of sepsis using electronic health record (EHR) data. They used retrospective EHR data from multiple Danish hospitals over seven years, including intensive care units and other departments. The data contained time-stamped events like medications, lab tests, notes, etc. They represented each event as a sparse vector. They developed a CNN-LSTM model, with CNN layers to extract features from event sequences and an LSTM layer to capture temporal dependencies. They compared to baseline models like gradient boosting on vital signs (GB-Vital) and a multilayer perceptron (MLP) model. The CNN-LSTM model achieved an AUROC of 0.856 at 3 hours before sepsis onset, outperforming the baselines. AUROC was 0.756 even 24 hours before onset. They proposed a new "sequence evaluation" approach to assess model utility over multiple predictions during a hospital stay. They also evaluated model utility by determining how many optimistic predictions were already on antibiotics/had blood cultures to see how much earlier the model could have helped. In summary, they developed a sequential deep learning model for early sepsis detection that outperformed standard models and proposed new evaluation approaches to assess clinical utility. The model could be generalized across different hospital departments.

Wardi et al. [25] developed and validated a machine learning model called the Artificial Intelligence Sepsis Expert (AISE) to predict the progression of septic shock in emergency department (ED) patients. They used data from >180,000 patients at two academic medical centers from 2014-2019. They identified patients with sepsis (using CMS and Sepsis-3 criteria) and modeled the risk of developing septic shock ≥4 hours after ED arrival. The AISE model uses 40 input variables and a neural network-integrated with a Weibull-Cox model to predict hourly risk. For CMS sepsis patients, AUC was >0.8 up to 12 hours before the shock, and AUROC was 0.833 at 12 hours in internal validation. They validated the model at a second site, using transfer learning to fine-tune it, significantly improving AUC to 0.85. The most predictive variables were vital signs like blood pressure, respiratory rate, and temperature. Mortality was 24.6% for delayed septic shock patients vs 5.5% for sepsis without shock, highlighting the importance of early prediction. In summary, they developed and validated an accurate machine learning model for predicting septic shock progression in ED patients that generalized well to a second site via transfer learning. The model could help identify high-risk patients early for intervention.

Ghias et al. [26] developed and compared machine learning models for early sepsis prediction in ICU patients. They used the PhysioNet 2019 challenge dataset, which has hourly measurements for 40,336 ICU patients from 2 hospitals. Several ML models, including XGBoost, Random Forest, LightGBM, Linear Learner, and Multilayer Perceptron, were trained and evaluated. Only six vital sign variables were used as features, selected based on statistical analysis. Missing data was imputed using MissForest. Based on the sepsis labels, the models were trained to predict sepsis 6 hours before onset. XGBoost had the best performance with an AUC of 0.98, accuracy of 0.98, precision of 0.97, and recall of 0.98. Random Forest and LightGBM also performed well. The ensemble models outperformed the baseline models. They analyzed model interpretation to understand the importance of features. Age, gender, and vital signs were most predictive of sepsis. In summary, they developed ML models for early sepsis prediction in ICU patients, with XGBoost showing the best predictive performance on the PhysioNet dataset based on vital signs and demographics. The models could help enable timely treatment.

Barghi and Azadeh-Fard [27] developed and compared machine learning models for predicting sepsis risk using early admission data. They used data from a hospital in Virginia with 20,005 patient records. About 7.5% had sepsis. The input features included demographics, severity level, mortality risk, admission type, and length of stay. Six ML models were developed: Logistic Regression, Naive Bayes, SVM, Boosted Tree, CART, and Bootstrap Forest. Models were evaluated on accuracy, F1-score, AUC, etc. Bootstrap Forest had the best AUC (0.908) and R-squared (0.366). Naive Bayes had the best F1-score (0.468), indicating good performance in the imbalanced classes. SVM and Boosted Tree had the lowest misclassification rate (0.063). All models had AUC > 0.88. Analysis showed higher sepsis rates for patients with higher severity levels, emergency admission, and age >80. In summary, they developed and evaluated multiple ML models for early sepsis risk prediction using admission data, with ensemble methods like Bootstrap Forest performing the best overall. The models could help enable timely treatment.

Table 1 presents a brief review of these studies:

*Table.1 A Brief Review of Previous Studies*

| Author | Year | Dataset | ML Models Used | Performance Metrics | Key Results |
|---|---|---|---|---|---|
| Abromavičius et al. | 2019 | PhysioNet Challenge (ICU) | Decision Tree, SVM, Naive Bayes, Ensemble | Utility score, AUCROC, Precision | Staged modeling based on ICU stay improved utility score to 0.245. Adjusting classification cost handled imbalance. Utility score was best metric. |
| Barton et al. | 2019 | UCSF, BIDMC (ICU) | Gradient Boosted Decision Tree | AUROC | AUROC 0.88 at onset using just 6 vital signs. Predicted up to 48 hrs prior. Outperformed existing scores. |
| Kaji et al. | 2020 | Large ICU dataset | RNN with LSTM & Attention | AUC | AUC 0.876 for sepsis. Attention provided interpretability. |
| Bloch et al. | 2021 | Rabin Medical ICU | Logistic Regression, SVM, ANN | AUC, Sensitivity, Specificity | SVM-RBF achieved AUC 88.38%. Focused on vital sign variability. |
| Liu et al. | 2022 | ICU, eICU | Random Forest, NN, POMDP | Precision, False alarms | Two-stage ML+POMDP improved precision by 9% & reduced false alarms by 28% |

| Author | Year | Dataset | ML Models Used | Performance Metrics | Key Results |
|---|---|---|---|---|---|
| Lauritsen et al. | 2020 | Multiple Danish hospitals | CNN-LSTM | AUROC, Precision, Sensitivity | AUROC 0.856 at 3 hrs before onset. Proposed SERAIP metric for clinical utility. |
| Wardi et al. | 2021 | 2 US hospitals | AI Sepsis Expert model | AUROC | AUROC >0.8 up to 12 hrs before delayed shock onset. Used transfer learning. |
| Ghias et al. | 2022 | PhysioNet challenge data | Xgboost, Random Forest, etc | AUC, Accuracy | Xgboost achieved AUC 0.98 & accuracy 0.98. Top predictors were vital signs. |
| Barghi and Azadeh-Fard | 2022 | Hospital in Virginia | Logistic Regression, Naive Bayes, SVM, etc | AUC, Accuracy, R-squared | Bootstrap Forest model had highest AUC of 0.908 and R-squared of 0.366. |
| Kausch et al. | 2021 | Systematic review | Various models | AUC | Model AUC ranged from 0.61 to 0.96. Highlights need for model validation standards. |

To sum up, prior studies have employed various modeling techniques to predict sepsis, each with its strengths and limitations. Our research aims to contribute to this field by comparing base models with the meta-ensemble model. The following sections of this paper will delve into the methodology, experimental setup, and results of our sepsis prediction model, offering a fresh perspective on this critical healthcare challenge.

**Methodology**

This This study uses an analytically applied methodology. Based on data from adult patients who attended the emergency department's electronic health record (EHR), we used machine learning algorithms to create models for identifying early sepsis risk. Vital indicators from the triage data were employed as the fundamental predictor variables. Our investigation used a random 80/20 data split into training and testing sets. Three machine learning algorithms—Random Forest, Decision Tree, and XGBoost—were used to construct models using the training set.

The Beth Israel Deaconess Medical Center (Hospital System A) and University of Amouri Hospital (Hospital System B) are two hospital systems in the United States with different EHR systems that are geographically distinct. The necessary institutional review board approvals were obtained before collecting this data throughout the previous ten years.

For data analysis, Google Colab notebooks employed the Python programming language. Open-source machine learning libraries for classification, clustering, regression, and dimensionality reduction applications include sci-kit-learn, numpy, pandas, and matplotlib. Popular library Scikit-learn is frequently used for feature extraction and model validation.

The time-series dataset used in this study has 44 columns. The data seems to be highly dispersed at first glance. The Beth Israel Deaconess Medical Center (Hospital System A), University of Amouri Hospital (Hospital System B), and an unknown hospital system (Hospital System C) were the three geographically dissimilar U.S. hospital systems from which data were gathered.

The entry and exit criteria for the study: Entry Requirements: Data were classified according to sepsis-3 clinical standards [7, 16, 17]. Public downloads of medical records for 40,336 patients from Hospital Systems A and B are accessible. Exit Criteria: This comprises variables with equal medical functionality, missing values, and unusual outliers in the dataset under examination.

The information used in this study was a compilation of time-series summaries of vital signs, laboratory results, and fixed patient characteristics. Forty clinical factors, including eight vital sign variables, 26 laboratory variables, and six demographic variables, were explicitly included in the data.

Before analysis and model creation, several preprocessing techniques were applied to the data retrieved from EHRs. Hourly windows were created by combining all patient features, which made developing and testing models easier. As an illustration, the average heart rate was calculated from several readings taken over an hour. Multiple Logical Observation Identifiers Names and Codes (LOINC) described the same clinical parameter, which were combined into a single compact variable. For instance, serum and arterial hemoglobin were combined into a single hemoglobin variable. The dataset used in this study exhibited a severe class imbalance, with a limited number of sepsis cases compared to non-sepsis patients. This class imbalance can lead to biased model performance. To mitigate this issue, we applied undersampling to balance the class distribution. Finally, after removing data, 18 variables with a high percentage of missing values and variables with identical medical functions were left. We constructed predictive models based on these 18 features.

Table 2 provides a list of variables used in the study.

*Table 2. List of variabales.*

|    | Variables | Categorical/numeric |
|----|-----------|---------------------|
| 1  | Hour      | numeric             |
| 2  | HR        | numeric             |
| 3  | O2Sat     | numeric             |
| 4  | Temp      | numeric             |
| 5  | MAP       | numeric             |
| 6  | Resp      | numeric             |
| 7  | BUN       | numeric             |
| 8  | Chloride  | numeric             |
| 9  | Creatine  | numeric             |
| 10 | Glucose   | numeric             |
| 11 | Hct       | numeric             |
| 12 | Hgb       | numeric             |
| 13 | WBC | numeric |
| 14 | Platelets | numeric |
| 15 | Age | numeric |
| 16 | HospAdTime | numeric |
| 17 | ICULOS | numeric |
| 18 | Sepsis Label | numeric |

Findings The entire medical dataset was randomly split into two subgroups in an 80:20 ratio, each labeled as the training and testing/validation subsets, using the train_test_split tool provided by the scikit-learn module.

Predictive base models were built using three algorithms: Random Forest, Decision Tree, and XGBoost. Our meta ensemble method is designed to combine the predictions of these three models to enhance overall performance. This involves the following steps:

a. Individual Model Training: Each base model (Random Forest, XGBoost, and Decision Tree) is independently trained on the dataset using appropriate hyperparameters and cross-validation techniques.

b. Meta Ensemble Creation: We create a meta ensemble by combining the predictions from the base models. This combination can be achieved through techniques such as simple averaging, weighted averaging, or more advanced methods like stacking or blending.

c. Ensemble Evaluation: The meta ensemble's performance is evaluated using appropriate evaluation metrics, such as Accuracy, Precision, F1-score, or Area Under the ROC curve (AUC).

Table 3 provides a demonstration of how well the sophisticated predictive models utilized in this study performed.

*Table 3. Performance of machine learning algorithms on 20% of the data.*

| Algorithm | Accuracy | Precision | Recall | F1 Score | AUC-ROC |
|---|---|---|---|---|---|
| Random Forest | 96% | 93% | 96% | 94% | 95% |
| XGBoost | 86% | 83% | 74% | 79% | 94% |
| Decision Tree | 89% | 80% | 89% | 84% | 90% |
| Meta-Ensemble Model | 95% | 94% | 96% | 95% | 96% |

**Feature Importance Analysis**

We indicate the feature importance analysis findings for the Random Forest model in this section. Understanding the variables that affect our model's prediction performance and learning about the underlying data require doing a feature importance analysis, which is a critical step.

## 1.1. Random Forest Feature Importance

Figure 1 illustrates the feature importance plot for the Random Forest model. The plot displays the relative importance of each feature in terms of its contribution to the model's predictive performance. Features are ranked in descending order of importance, with the most essential features on the plot's left side.

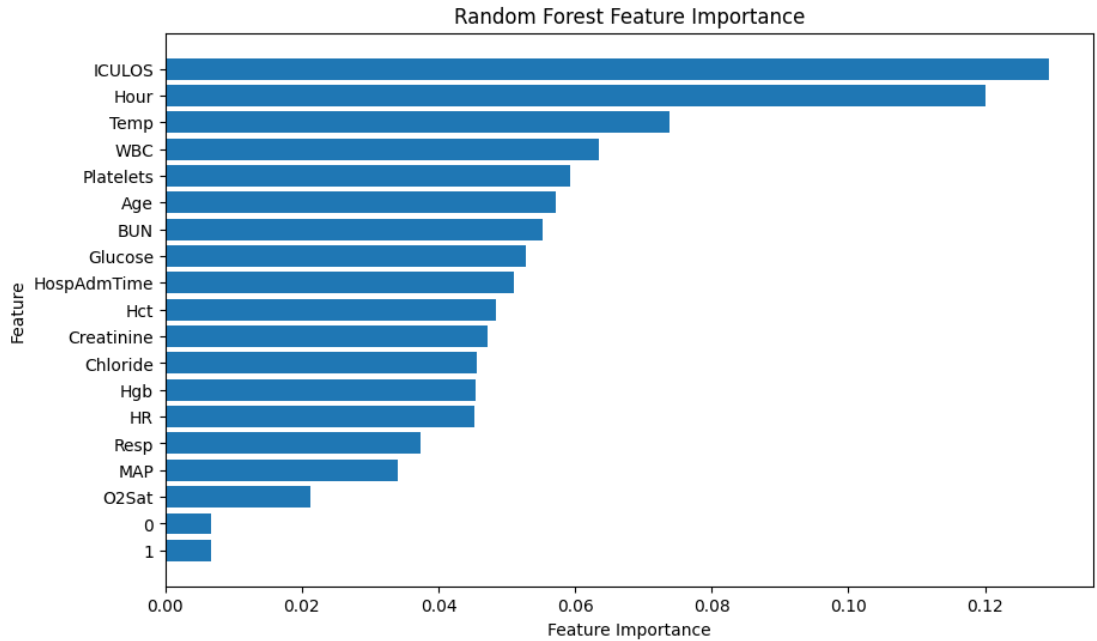

*Figure 1: Random Forest Feature Importance Plot*

As observed in Figure 1, ICULOS, Hour, and Temp are the top three most important features, with ICULOS having the highest importance score. This suggests that these features play a significant role in predicting the target variable.

## 1.2. Interpretation and Insights

The feature importance analysis reveals several noteworthy insights:

*ICULOS:* This feature demonstrates the highest importance score, indicating that it has the most substantial influence on the model's predictions. Further investigation into the nature of ICULOS may provide valuable insights into the underlying mechanisms of the problem.

*Hour and Temp:* Both Feature C and Feature D also exhibit high importance scores, suggesting their significant contribution to the model's predictive power. Understanding the domain relevance and context of these features could lead to actionable recommendations.

*Other Features:* While the top features mentioned above are the most influential, it is essential to note that other features also contribute to the model's performance. Even seemingly less important features may contain valuable information, and their collective contribution improves the overall predictive accuracy of the model.

### 1.3. Decision Tree Feature Importance

Figure 2 illustrates the feature importance plot for the Decision Tree model. Unlike ensemble models like Random Forest, Decision Trees calculate feature importance differently. In this case, the importance is computed based on the Gini impurity reduction associated with each feature.

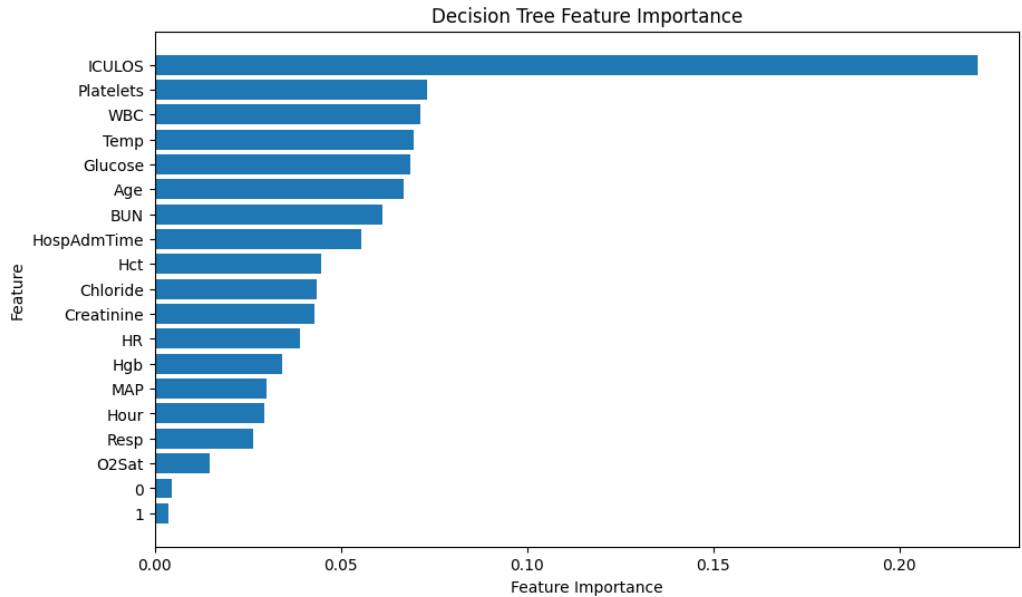

*Figure 2: Decision Tree Feature Importance Plot*

As shown in Figure 2, ICULOS exhibits the highest Gini impurity reduction, making it the most important featur for the Decision Tree model. This suggests that this features has a significant impact on the model's ability to partition the data effectively.

### 1.4. XGBoost Feature Importance

Figure 3 illustrates the feature importance plot for the XGBoost model. XGBoost assigns feature importance scores based on the number of times each feature is used for splitting across all trees in the ensemble.

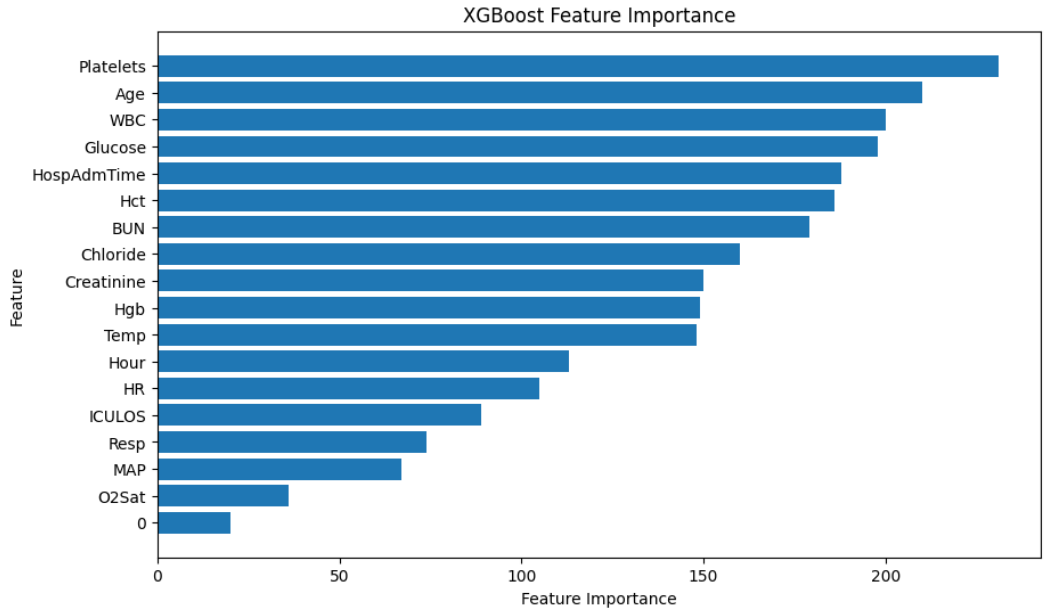

*Figure 3: XGBoost Feature Importance Plot*

As depicted in Figure 3, Platelets, Age, and WBC are the top three most important features in the XGBoost model, with Age having the highest importance score. This implies that these features play a pivotal role in making predictions.

*Interpretation and Insights*

The feature importance analysis for the XGBoost model provides the following insights:

Platelets: This feature has the highest importance score, indicating its substantial influence on the model's predictions. Further analysis of Age's relationship with the target variable may provide valuable insights.

Age and WBC: Both Platelets and WBC also exhibit high importance scores, signifying their significant contribution to the model's predictive power.

Other Features: While the top features mentioned above are the most influential, the model also benefits from the collective contribution of other features.

 *Importance Analysis of Base Models*

In this section, we present an analysis of the importance of the three base models utilized in our meta-ensemble: Random Forest, XGBoost, and Decision Tree. The objective is to understand the relative contributions of these models to the overall predictive power of the ensemble.

*Importance Scores of Base Models*

To assess the importance of each base model, we conducted an importance analysis and visualized the results in Figure 4:

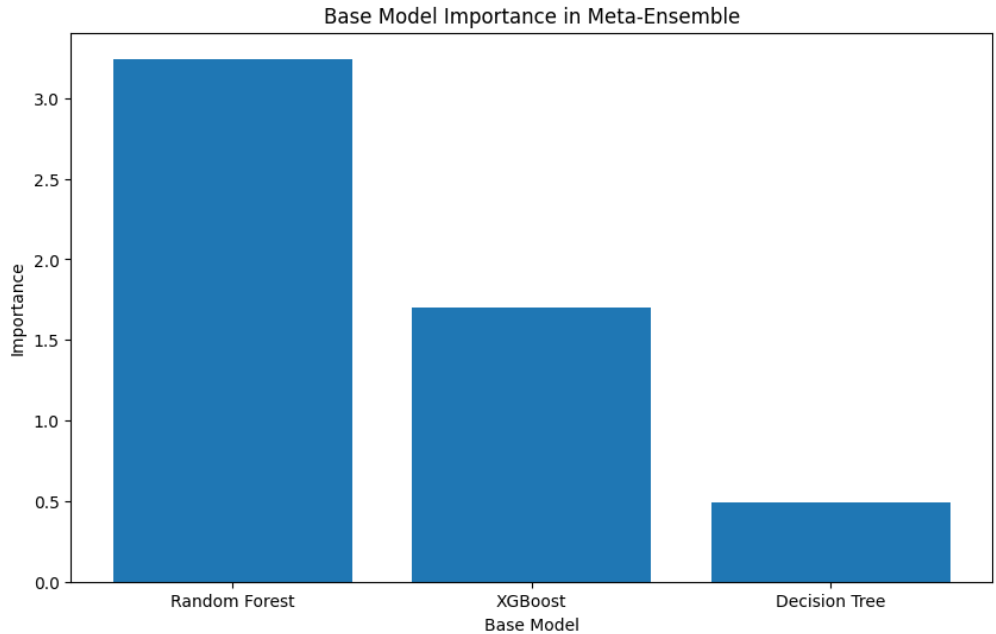

*Figure 4: Importance Scores of Base Models*

Figure 4. illustrates the importance scores of the base models within our meta-ensemble. The scores were calculated based on their impact on the final ensemble's predictive performance.

As depicted in Figure 4, the importance scores are as follows:

*Random Forest:* This base model exhibits the highest importance score among the three. Its ability to handle complex relationships in the data and reduce overfitting contributes significantly to the ensemble's performance.

*XGBoost:* The XGBoost base model follows closely in importance. Its gradient boosting techniques enhance predictive accuracy and provide a substantial boost to the ensemble's overall performance.

*Decision Tree:* While Decision Tree has the lowest importance score in the ensemble, it still contributes positively. The simplicity and interpretability of decision trees make them valuable components, even when their importance is relatively lower.

**Results**

In this section, we present the results of our study, which focused on predicting sepsis using four distinct machine learning models, namely Random Forest, XGBoost, Decision Tree, and a Meta-Ensemble model. We evaluate the performance of these models using key metrics such as Accuracy, Precision, Recall, F1 score, and the Area Under the Receiver Operating Characteristic Curve (AUC-ROC). These metrics provide valuable insights into the models' classification accuracy, precision in positive predictions, ability to capture true positives, and overall discriminative power.

*Accuracy:* The Random Forest model has the highest accuracy, indicating it correctly predicts the most instances. The Meta-Ensemble follows closely behind. XGBoost has the lowest accuracy among these models.

*Precision:* Random Forest and the Meta-Ensemble have the highest precision, meaning they make fewer false positive predictions. XGBoost has slightly lower precision, while Decision Tree has the lowest precision.

Recall: Random Forest and the Meta-Ensemble have the highest recall, indicating that they capture more of the positive cases. XGBoost has a lower recall, and Decision Tree has the second-highest recall.

*F1 Score:* The F1 Score is the harmonic mean of precision and recall, balancing false positives and false negatives. The Meta-Ensemble has the highest F1 score, indicating a good balance between precision and recall. Random Forest follows closely, while XGBoost has the lowest F1 score.

*AUC-ROC:* The AUC-ROC measures the model's ability to distinguish between positive and negative classes. In the context of early sepsis prediction, we developed three distinct machine learning models: Random Forest, XGBoost, and Decision Tree. Each of these models exhibited varying levels of predictive performance as measured by the Area Under the Receiver Operating Characteristic Curve (AUC-ROC). Specifically, the AUC-ROC scores for the individual models are as follows:

Random Forest: AUC = 0.95, XGBoost: AUC = 0.94, Decision Tree: AUC = 0.90

These results reflect the ability of each base model to discriminate between sepsis and non-sepsis cases based on the given features. To further enhance our predictive capabilities, we employed a meta-ensemble approach that leverages the predictions of the base models. This meta-ensemble model yielded an AUC-ROC score of 0.96, surpassing the performance of each individual base model. The Receiver Operating Characteristic (ROC) curves visually illustrate the discriminative power of each model. As shown in the plot below, the Meta-Ensemble model's ROC curve lies closer to the top-left corner, indicating superior overall performance compared to the base models:

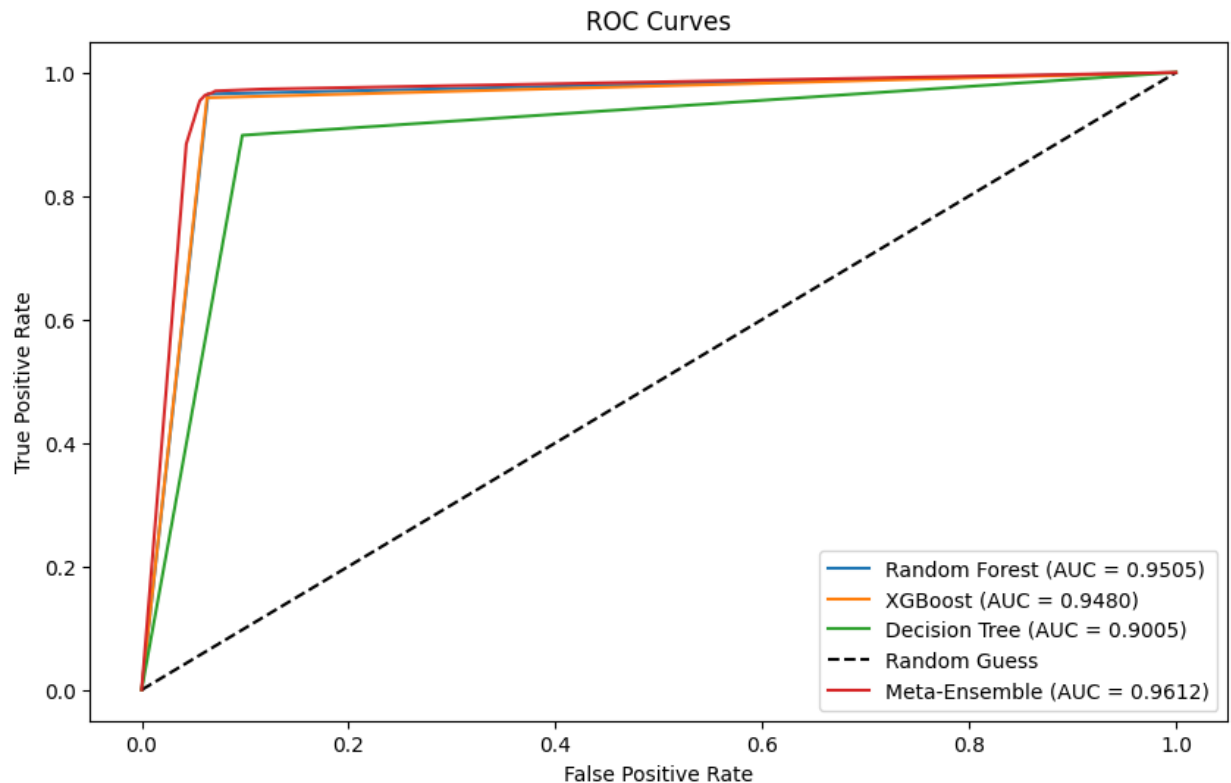

The notable improvement in AUC-ROC score observed in the Meta-Ensemble model demonstrates its effectiveness in early sepsis prediction, consolidating information from the base models to achieve enhanced predictive accuracy. All in all the Random Forest and the Meta-Ensemble models perform exceptionally well across most metrics, with high accuracy, precision, recall, and F1 score. The Meta-Ensemble combines the strengths of the three base models to achieve the best overall performance. XGBoost also performs well but has slightly lower accuracy and F1 score compared to the Meta-Ensemble. The Decision Tree model, while not as strong as the others, still demonstrates respectable performance in some metrics. These findings suggest that the meta-ensemble approach successfully leverages the strengths of multiple base models to improve predictive performance in the context of early sepsis detection.

**Discussion and Conclusion**

The results of this study reinforce the urgency of improving sepsis detection and underscore the potential of machine learning techniques in addressing this critical healthcare challenge. Early prediction of sepsis is essential, as delays in intervention correlate with higher mortality rates. Despite their widespread use, existing clinical screening tools have demonstrated limitations in sepsis detection. The machine learning models employed in this study, including Random Forest, XGBoost, and Decision Tree, showcased their ability to discriminate between sepsis and non-sepsis cases using a variety of clinical variables. Their individual AUC-ROC scores highlighted their potential for sepsis prediction, with the Random Forest model leading in performance.

However, our research's standout innovation lies in applying a meta-ensemble approach to combine predictions from these base models. This approach significantly improved predictive accuracy. The meta-ensemble model exhibited a higher AUC-ROC score than the individual base models, indicating its superior overall discriminative power. One of the notable findings is that the meta-ensemble model excelled in recall, indicating its proficiency in identifying a substantial portion of actual positive cases while maintaining competitive accuracy and precision. This characteristic makes it well-suited for early sepsis prediction. These results underscore the potential of a meta-ensemble approach as a promising strategy for enhancing sepsis prediction accuracy. By amalgamating predictions from various base models, we leveraged the strengths of each, addressing their limitations. This research contributes to the growing body of literature on sepsis prediction and offers a fresh perspective on tackling this critical healthcare challenge.